\documentclass{article}

\usepackage{PRIMEarxiv}

\usepackage[utf8]{inputenc} 
\usepackage[T1]{fontenc}    
\usepackage{hyperref}       
\usepackage{url}            
\usepackage{booktabs}       
\usepackage{amsfonts}       
\usepackage{nicefrac}       
\usepackage{microtype}      
\usepackage{lipsum}
\usepackage{fancyhdr}       
\usepackage{graphicx}       
\graphicspath{{media/}}     

\usepackage{adjustbox}
\usepackage{bbm}
\usepackage{bm}
\usepackage{amsmath}
\usepackage{booktabs}

\usepackage{makecell}
\usepackage{caption}
\usepackage{subcaption}
\usepackage{multirow}
\usepackage{xurl}
\usepackage[dvipsnames]{xcolor}

\usepackage{microtype}

\usepackage{array}
\usepackage{enumitem}
\usepackage{bm}
\usepackage{amsmath}
\usepackage{booktabs}
\usepackage{algorithm}
\usepackage{algpseudocode}
\usepackage{amssymb}
\usepackage{pifont}
\usepackage{tabularx}
\newcolumntype{Y}{>{\centering\arraybackslash}X}

\title{Bias and Fairness in Chatbots: An Overview}

\usepackage{authblk}
\author[1]{Jintang Xue}
\author[1]{Yun-Cheng Wang}
\author[1]{Chengwei Wei}
\author[2]{Xiaofeng Liu}
\author[2]{Jonghye Woo}
\author[1]{C.-C. Jay Kuo}
\affil[1]{University of Southern California, Los Angeles, California, USA}
\affil[2]{Massachusetts General Hospital and Harvard Medical School, Boston, MA, 02114, USA}

\begin{document}
\maketitle

\begin{abstract}

Chatbots have been studied for more than half a century. With the rapid
development of natural language processing (NLP) technologies in recent years,
chatbots using large language models (LLMs) have received much attention
nowadays.  Compared with traditional ones, modern chatbots are more
powerful and have been used in real-world applications.  There are
however, bias and fairness concerns in modern chatbot design. 
Due to the huge amounts of training data, extremely
large model sizes, and lack of interpretability, bias
mitigation and fairness preservation of modern chatbots are challenging. 
Thus, a comprehensive overview on bias and fairness in chatbot systems 
is given in this paper.
The history of chatbots and their categories are first reviewed.  Then,
bias sources and potential harms in applications are analyzed. Considerations in
designing fair and unbiased chatbot systems are examined. Finally,
future research directions are discussed. 
    
\end{abstract}

\section{Introduction}

A chatbot is an intelligent software system designed to simulate natural
human language conversations between humans and machines 
\cite{caldarini2022literature}. As a human-computer interaction (HCI) system
\cite{chaves2021should}, it takes human voice or text as input and uses
the natural language processing (NLP) technology to understand and
respond accordingly \cite{adamopoulou2020overview}.  With the rapid
development of the Internet and artificial intelligence (AI), chatbots
have become a hot research topic and a real-world application system
that attracts much attention \cite{nirala2022survey}.  

One of the most common occasions is to use chatbots as a dialogue agent
in the service industry \cite{suhaili2021service, adam2021ai,
lalwani2018implementation}.  Chatbots have changed the way customers and
companies interact. While chatbots may not be as good as human services
in answering complex questions, they are accessible, responsive, and always
available. They can answer most simple questions, which proves to be
valuable in applications like product ordering and travel booking
\cite{kaczorowska2019chatbots, luo2019frontiers, ukpabi2019chatbot}. For
companies, chatbots can respond to customer requests at any time,
improve user experience, and contribute to saving in the service cost
\cite{zumstein2017chatbots}. As to users, a study \cite{brandtzaeg2017people}
showed that people would be interested in
chatbot services for effective and efficient information access.  Other
motivations include entertainment, socializing, and curiosity about new
things. To realize these benefits, chatbots need to understand user
input and analyze users' sentiments and intentions accurately, find
appropriate answers, and generate fast and fluent responses. Sometimes,
it may need to take the user identity (or attributes) into account in
providing a proper answer. 

Recent advances and breakthroughs in NLP and machine learning (ML) have
changed the landscape of language understanding and processing
\cite{otter2020survey, khurana2023natural, wei2023overview}.  These
developments are driven by the availability of increased computing
power, massive amounts of training data, and the advent of sophisticated
ML algorithms. The introduction of transformer networks
\cite{vaswani2017attention} leads to large pre-trained models, such as
GPT-3 \cite{brown2020language}, BERT \cite{devlin2018bert}, PaLM
\cite{chowdhery2022palm}, etc. They have become popular
\cite{qiu2020pre, kalyan2021ammus, wang2023large, zhao2023survey} in the
past decade.  Based on these developments, ChatGPT, a chatbot from
OpenAI, has taken the world by storm by providing real-time,
plausible-looking responses to input questions.  ChatGPT has a good
performance in text generation, language understanding, and translation.
As a chatbot, it can be applied in various fields
\cite{fraiwan2023review}, such as education \cite{firat2023chat,
baidoo2023education}, healthcare \cite{biswas2023role,
sallam2023chatgpt}, marketing \cite{jain2023prospects,
rivas2023marketing}, environmental research \cite{zhu2023chatgpt,
biswas2023potential}, etc. The prevalence of ChatGPT has made chatbots a
focus of attention. Leading technology companies have also released
their own chatbots, such as Google's Bard and Meta's BlenderBot 3. 

\begin{figure}[t]
\centering
\includegraphics[width=0.8\textwidth]{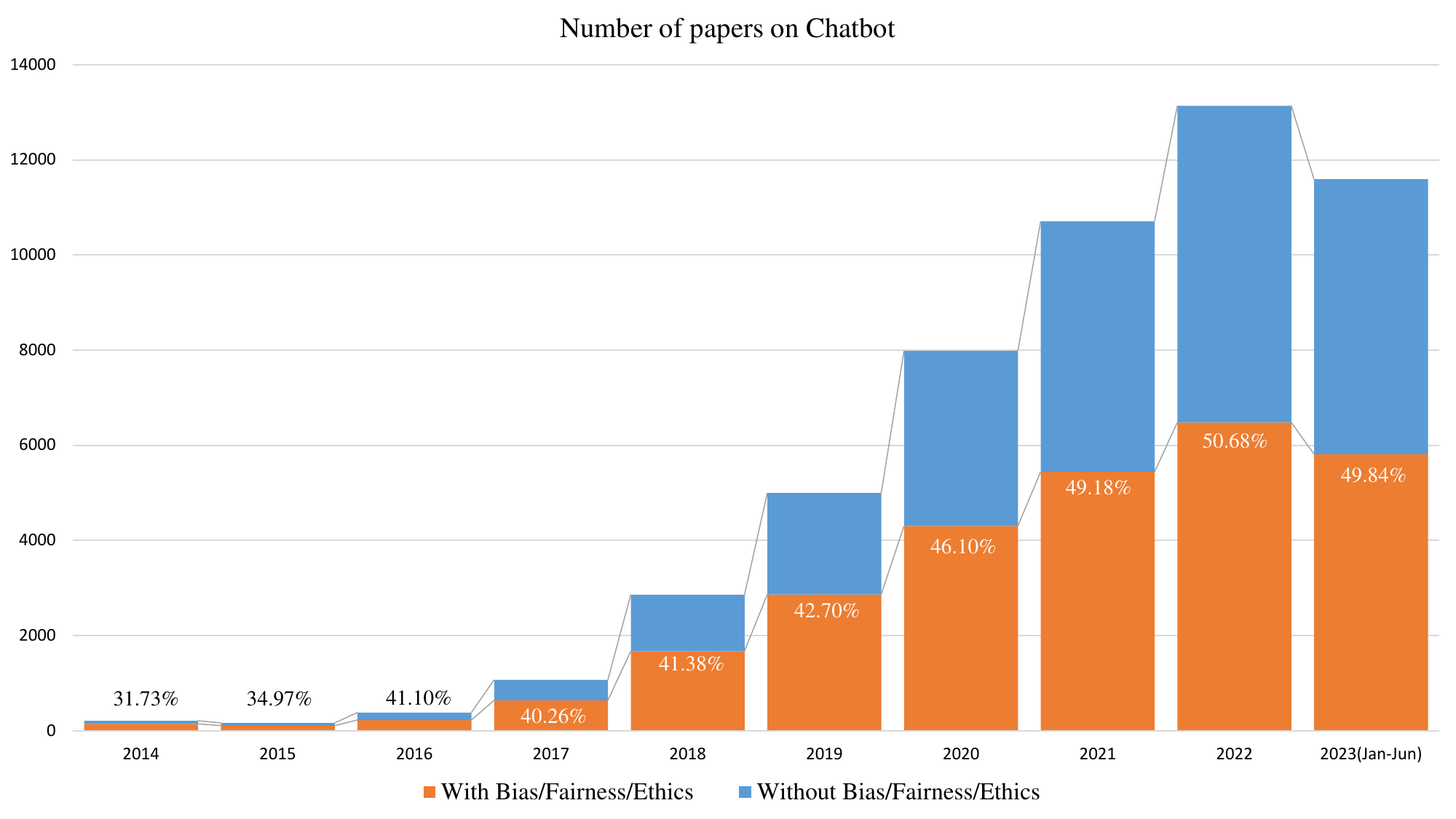}
\caption{Search results by the year for ``chatbot" or ``ChatGPT" as
keywords in the website of ``Dimensions" since 2014.}
\label{fig:number_of_papers}
\end{figure}

With the help of AI, chatbots have become more intelligent and can
answer people's questions smoothly. On the other hand, chatbots are not
as neutral as expected, raising ethical concerns among the general
public \cite{paul2023chatgpt}. Fig.~\ref{fig:number_of_papers} shows the number
of papers on chatbot
since 2014.  The number of papers on chatbots has risen sharply since
2015.  Noticeably, the number of papers in the first half of 2023 has
exceeded that in 2021.  All of them provide strong evidence of people's
attention to chatbots.  Furthermore, we see from the figure that about
one-half of them talk about bias, fairness, or ethics every year.
It suggests that, as chatbots become more advanced, concerns about
ethical issues also increase. Since the launch of ChatGPT, many papers
have been published on this topic.  Some analyze
its bias and fairness in general \cite{ferrara2023should,
ray2023chatgpt} while others are concerned with the same problem in
specific applications. For example, ChatGPT may have political bias
\cite{rozado2023political, rutinowski2023self}, bias against
conservatives \cite{mcgee2023chat}, bias in healthcare and education
\cite{sallam2023chatgpt, hosseini2023exploratory}, etc. The power of LLMs
has spawned many modern chatbots, and ChatGPT is only one of them.
Although there are papers on bias in general chatbots, they only examine
a narrow aspect, such as gender bias \cite{feine2020gender} and
stereotypes \cite{lee2019exploring}. They do not examine bias sources in
chatbot applications in our society systematically. This is the void
that we attempt to fill in this overview paper.

Although ChatGPT brings the ethical concerns of chatbots into the
spotlight, bias in chatbots is actually not a new topic
\cite{schlesinger2018let}. Most of the existing well-performing language
models are ML-based models. ML algorithms face the bias problem in many
aspects, such as data, user interaction, the algorithm itself, etc.
\cite{mehrabi2021survey, suresh2019framework}. As a special ML system
that interacts with humans directly, chatbots have a greater impact on
ethical issues. To give an example, Microsoft released an AI-based
chatbot called Tay via Twitter in 2016. It had the ability to learn from
conversations with Twitter users.  However, data obtained from Twitter
users were seriously biased. Shortly after the chatbot was released, its
speech turned from friendly, kind speech to discriminatory, offensive,
and inflammatory speech in a short time. As a result, Microsoft had to
shut down the chatbot urgently within a day after releasing it
\cite{neff2016talking, wolf2017we}.  Similar risks exist in recent
chatbots. OpenAI CEO, Sam Altman, admitted that they were aware of
ChatGPT's shortcomings in terms of bias in a Twitter thread in February
2023. Later, he added that technologies could ``go quite wrong'' and his
``biggest fear'' was that they would cause significant harm to the
world. Some people with ulterior motives may take advantage of the flaws
in chatbots and use them to harm the society.  To alleviate these
problems, government regulation could be effective \cite{chan2023gpt,
lagrandeur2021safe, burtell2023artificial}.  However, finding a balance
between regulation and freedom of use is a problem that remains to be
investigated, as over-regulation can hinder the development of
innovation \cite{zhou2023ethical}. 

Ethical issues can be a barrier for companies to use large language
models (LLMs) to interact with customers. In particular, the use of
black-box models that lack transparency and interpretability to
communicate with users is dangerous and unpredictable. A good chatbot
can improve the user experience on the original basis, while a biased
chatbot can cause a devastating blow to the user experience and cause
serious damage. Recently, there are quite a few papers talking about the
bias and fairness issues of ML systems, NLP algorithms, and ChatGPT
applications.  In contrast, there are fewer papers on bias and fairness
in designing chatbot systems, which is the main focus of this overview
paper. The main contributions of our work include the following. 
\begin{itemize}
\item A comprehensive review of the history, technologies, and
recent developments of chatbots. 
\item Identification of bias sources and potential harms in chatbot 
applications. 
\item Considerations in designing fair chatbot systems and future
research topics.
\end{itemize}
The rest of the paper is organized as follows. The chatbot history, architectures, and
categories are examined in Sec.~\ref{sec: Chatbots}. Possible bias
sources, caused harms, and matigation methods in applications are discussed in Sec.~\ref{sec:
Bias}. Considerations in designing a fair chatbot system are presented
in Sec.~\ref{sec: Fairness}.  Future research directions are pointed out
in Sec.~\ref{sec: Future}.  Concluding remarks are given in Sec.
\ref{sec: Conclusion}. 

\section{History, Architectures, and Development Categories 
of Chatbots}\label{sec: Chatbots}
    
\subsection{History of Chatbots} 

\begin{figure}[htb]
\centering
\includegraphics[width=0.9\textwidth]{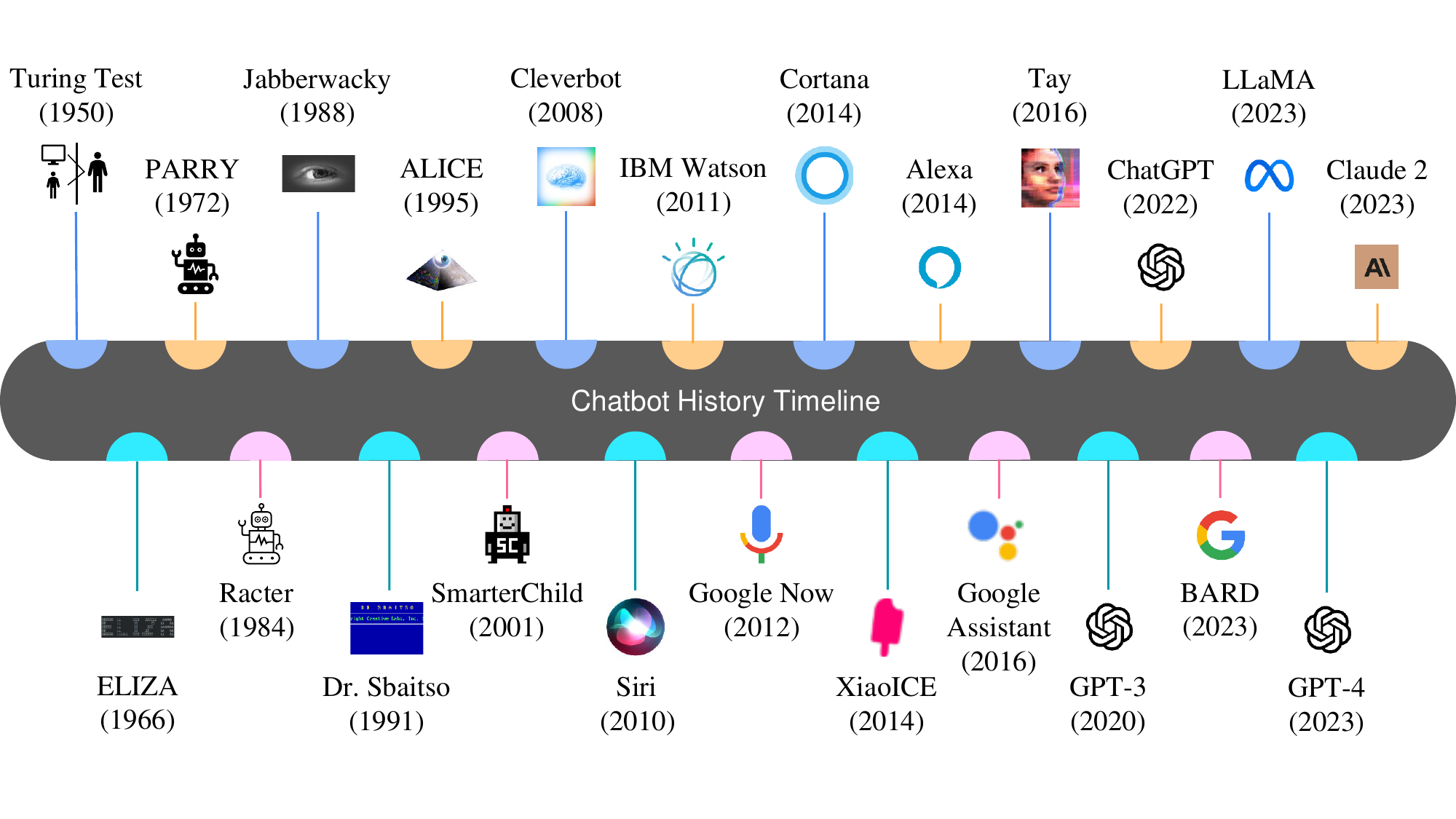}
\caption{The history of chatbots.}\label{fig:chatbot_history}
\end{figure}
    
The history of chatbots is depicted in Fig.~\ref{fig:chatbot_history}.
The concept of a chatbot was initiated in the Turing test in 1950.
Various forms of chatbots have evolved over five decades in stages.
Modern chatbots are built upon LLMs. The history of chatbots is briefly
described below. 
        
{\bf 1) Conceptual Stage.} To answer the question ``whether a machine can
think?'', Turing proposed the question-and-answer paradigm in 1950,
which is known as the Turing test \cite{turing2009computing}. Simply
speaking, one human participant would like to judge whether the other
participant is a machine or a person through text questions and answers
(rather than voice and/or appearance).  If the human participant can
hardly tell if the other participant is a machine or not, we may claim
such a machine can think.  The conversation idea between humans and
robots through text was conceived in such a test. This is viewed as the
origin of chatbots. 
        
{\bf 2) From 1960 to 1980.} An early well-known chatbot, ELIZA, was
developed by Weizenbaum at MIT to simulate a Rogerian psychotherapist in
1966 \cite{weizenbaum1966eliza}. It found keywords from the input, 
reassembled user input and pre-prepared responses through certain rules
to generate responses. ELIZA did not understand the meaning of the input
during the process. It just conducted pattern matching and substitution.
However, some of its responses made it difficult for people, who used
the program for the first time at the time, to tell whether it was a
machine or a human. Some people even developed an emotional attachment
to it.  The latter raised some ethical considerations
\cite{weizenbaum1976computer}.  Although ELIZA caused a sensation in
60s, it had many shortcomings.  For example, its knowledge base was
limited, and it could only answer questions in a certain narrow range.
On the other hand, its appearance played an important role in inspiring
follow-up research. Another famous early chatbot, PARRY, was developed
to simulate a person with paranoid schizophrenia in 1972
\cite{colby1971artificial}. PARRY interacted with ELIZA, who played the
role of Rogerian's therapist. PARRY was considered more advanced than
ELIZA because it had a better controlling structure and displayed some
emotions \cite{colby1981modeling, zemvcik2019brief}. 
        
{\bf 3) From 1980 to 2010.} More explorations of chatbots were made from
80s and 00s. Racter was an AI program released in 1984. It generated
prose in English, and its interactive version behaved like a chatbot. A
learning AI project, named Jabberwacky, was conducted in 1988.  It was
designed to simulate natural human chatting in a fun way
\cite{shawar2007fostering}. Different from earlier chatbots, Jabberwacky
learned from chatting with people and stored keywords in previous
conversations to grow its knowledge base \cite{singh2020survey}. Then,
it used context matching with a dynamically growing database to choose
appropriate responses \cite{kerlyl2006bringing}. Its new version, called
Cleverbot, was released in 2008. Creative Labs designed a chatbot, named
Dr. Sbaitso, for MS-DOS computers in 1991 and released it together with various sound
cards in the 1990s \cite{deryugina2010chatterbots}. Its interactive interface
was a blue background with a white font. Although the interactive
content was relatively simple, it used the speech synthesis technology
and the sound card to realize text-to-speech (TTS) in the early stage.
Inspired by ELIZA, Wallace developed ALICE (Artificial Language Internet
Computer Entity) \cite{wallace2009anatomy} in 1995. ALICE still used
pattern matching rules but it was more capable since it had a much
larger knowledge base. It used AIML (Artificial Intelligence Markup
Language) to specify chat rules. Specifically, it used categories as
basic knowledge units, where each category contains patterns and
templates as user inputs and the corresponding machine responses,
respectively \cite{shawar2002comparison}.  ALICE gained significant
recognition at the time. For example, it won the Loebner Prize three
times in 00s \cite{bradevsko2012survey}. However, it still failed the
Turing test due to some limitations \cite{shum2018eliza}. ActiveBuddy
developed a chatbot, named SmarterChild, on the AIM platform in 2001.
SmarterChild was one of the earlier chatbots that could help people with
daily tasks through interaction, such as checking weather conditions,
showtimes, stocks, etc.  \cite{adamopoulou2020chatbots}. 

{\bf 4) From 2010 to 2020.} Watson was developed by IBM as a
question-answering chatbot in 2011 \cite{high2012era}. It participated
in the ``Jeopardy'' quiz show and won the championship twice. It was
later used in the healthcare \cite{chen2016ibm}. Microsoft developed a
chatbot called XiaoICE \cite{zhou2020design} based on the emotional
computing framework in 2014.  It had both IQ and EQ modules and could
flexibly answer user's questions. It was deployed in multiple countries
and platforms.  Another chatbot called Tay was released by Microsoft via
Twitter in 2014. It learned from users but learned inappropriate remarks
very fast, forcing Microsoft to shut it down shortly. 

Furthermore, chatbots have been widely used in people's daily lives in
the form of voice/search agents in instant messaging devices
\cite{hoy2018alexa, kepuska2018next}.  Siri was released as an iOS app
in February 2010 and integrated into iOS in 2011.  It has been part of
Apple's products since then. As a personal assistant, Siri can accept
users' voice inputs and complete tasks such as making calls, reminding,
looking for information, and translating \cite{aron2011innovative}.
Google released Google Now as a Google voice search app in 2012. It
takes users' voices as input and returns with searched results.
Microsoft launched Cortana for its Windows operating system in 2014.  It
responds to users' inputs using the Bing search engine.  Amazon launched
Alexa, together with the Echo speaker, in 2014.  Google launched Google
Assistant and integrated it with Google Home speakers and Pixel
smartphones in 2016. These voice assistants connect to the Internet and
respond quickly. However, they face multilingual, privacy, and security
challenges \cite{bolton2021security}. 

{\bf 5) After 2020.} The advancement of LLMs has impacted the development of
chatbots greatly since 2020 \cite{wei2023overview,
zhou2023comprehensive}. The transformer-based NLP technologies have made
major breakthroughs in natural language understanding and generation
\cite{adiwardana2020towards, roller2020recipes, cao2023comprehensive}.
LLM-based chatbots can provide rich responses using extensive training
conducted on large pre-trained transformers. GPT-3 was released by OpenAI in 2020
and it laid the foundation of ChatGPT. ChatGPT was released in
2022 and gained more than 100 million users \cite{wu2023brief} shortly.
Unlike previous chatbots, ChatGPT is an open-domain chatbot that can
answer questions across a wide range of domains. LLMs have brought
chatbots to a new level. 

On the other hand, the popularity of ChatGPT has also led to a lot of
controversy.  As a large generative AI model, ChatGPT has a huge number
of parameters. Its responses are difficult to predict and control,
raising concerns about trustworthiness, toxicity, bias, etc.
\cite{zhuo2023exploring}. Its responses, which are highly similar to
human beings, have aroused severe concerns.  OpenAI released an even
larger and more powerful LLM, called GPT-4, in 2023. The emergence of
ChatGPT has impacts on the AI industry. In response to ChatGPT, Google
launched Bard, a conversational generative artificial intelligence
chatbot powered by LaMDA \cite{thoppilan2022lamda}, in 2023. Meta
announced its own LLM, called LLaMA \cite{touvron2023llama}. Anthropic
released Claude2. 

\subsection{Architectures of Chatbots}

\begin{figure}[t]
\centering
\includegraphics[width=0.8\textwidth]{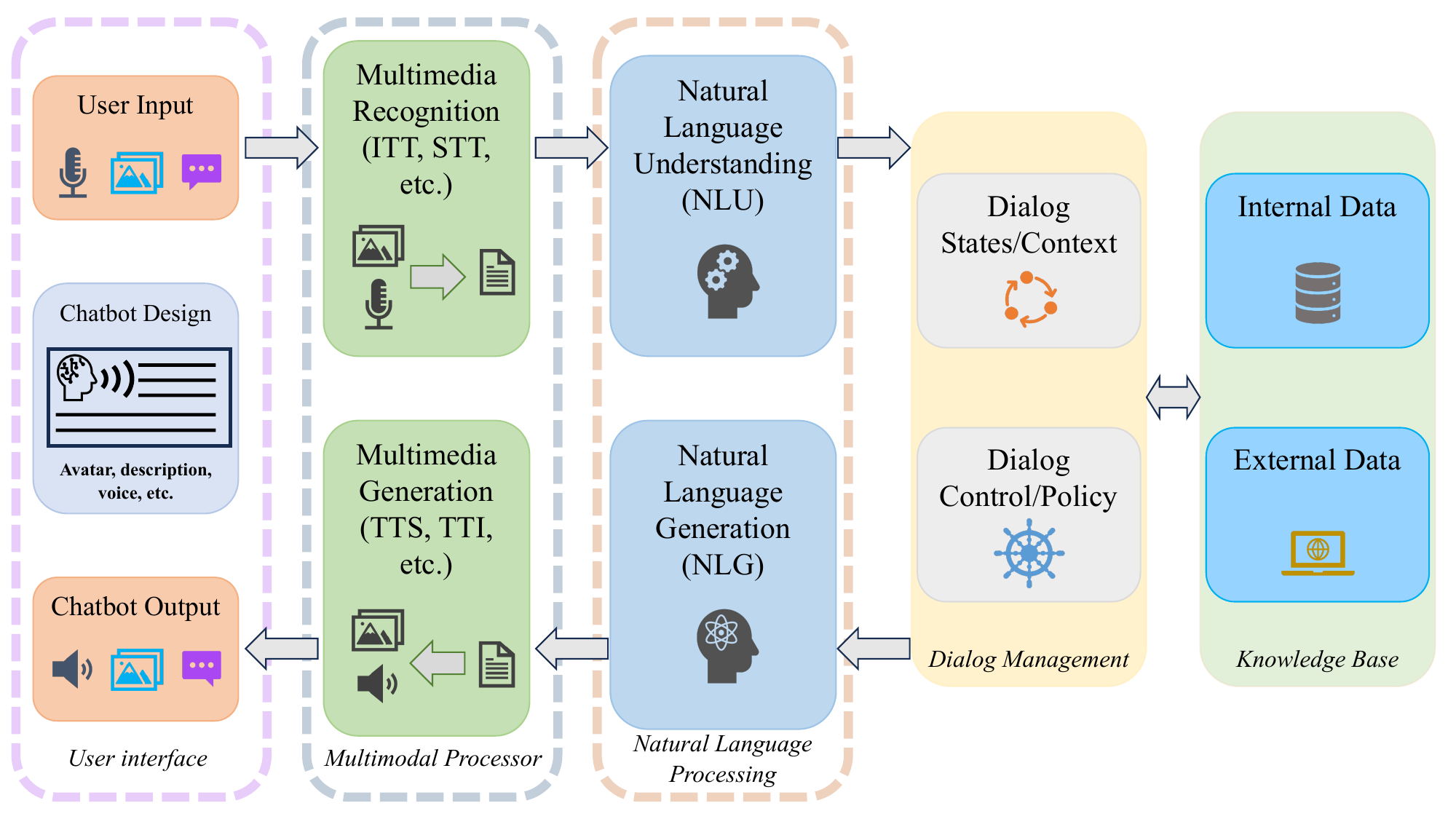}
\caption{The architecture of a general chatbot.}\label{fig:chatbot_arc}
\end{figure}

The architecture of a general chatbot is shown in
Fig.~\ref{fig:chatbot_arc}.  It consists of five main modules: 1) user
interface, 2) multimedia processor, 3) natural language processing, 4)
dialogue management, and 5) knowledge base.  The user interface module
is responsible for input and output of the chatbot. It is the module
that interacts with users directly. The input and output can be
multi-modal.  Multi-modal data is processed by the multimedia processor
module. The NLP module is used to understand user's input language and
generates the desired output language based on text answers. The
dialogue management module is responsible for recording the current chat
status and guiding the direction of conversations. It can access the
database module and get answers for users.  Because of the introduction
of end-to-end LLMs, boundaries between various modules may not be as
clear as those in the traditional chatbot design. Yet, there are still
sub-modules that are responsible for the above-mentioned functions. 
The roles of these five modules are elaborated below.

{\bf User Interface.} The user interface is a module that allows users
to interact with the chatbot. It receives input from the user and
provides output generated by the chatbot to the user. The system allows
input and output of multiple modalities such as text, voice, or even
pictures. Earlier chatbots used the text input. Voice assistants have
been common in various devices nowadays. Recently, a variety of
multimedia, such as images and videos, can also serve as input and output.
To make chatbots more realistic, additional features, such as chatbot
avatars, voices, emotions, etc., were added. These features can enhance
user experience, making human-machine interaction smoother
\cite{angga2015design, folstad2018makes}. For example, Microsoft's
chatbot, XiaoIce, appears as an 18-year-old girl in a Japanese school
uniform to users. She has a relatively complete resume and can do
self-introduction. All these additions make the chatbot easier to be
accepted as a virtual companion by users. 

{\bf Multimodal Processor.} In order to realize multimedia input and
output (rather than text alone), chatbots need to convert data from different
modalities to their corresponding text embeddings for further processing.  For example, for
speech input, the speech-to-text (STT) technique or the automatic speech
recognition (ASR) technique \cite{malik2021automatic, nassif2019speech}
can be used. For images, the image-to-text \cite{zelaszczyk2023cross,
li2019visual} (ITT) technique has been developed. On the other hand, in
response to users, chatbots need to convert text-embedded responses
generated by the chatbot to various modalities, such as text-to-speech
(TTS) or speech synthesis \cite{tan2021survey, ning2019review},
text-to-image (TTI) \cite{zhang2023text, frolov2021adversarial}, etc. In
recent years, these techniques have been greatly improved because of the
advancement of AI/ML \cite{baltruvsaitis2018multimodal}. 

{\bf Natural Language Processing.} Natural Language Understanding (NLU)
and Natural Language Generation (NLG) are two subtopics of Natural
Language Processing (NLP). They are both key components in chatbot
systems.  NLU takes human text as input and converts it into a form that
computers can understand and process. Two important tasks of NLU are
intent recognition and entity recognition \cite{jiao2020intelligent}.
Intent recognition refers to understanding user's intention and
observing user's emotion. It serves as the basis for chatbots to
generate reasonable answers. Irrelevant answers degrade user experience
significantly. Entity recognition refers to the extraction of entities
that exist in real life in user input sentences, such as objects,
people, cities, etc. These entities help the chatbot make logical
reasoning and find answers in the knowledge base. NLG is another
important component, which is responsible for generating human language
fluently and naturally \cite{gatt2018survey}. Using NLG, computers
generate emotional responses using human natural language, thereby
enhancing user experience and trust. 

{\bf Dialog Management.} Dialogue management (DM) is used to decide the
communication strategies \cite{cahn2017chatbot}. It needs to remember
the current dialogue state and control the content of the next dialogue.
After acquiring user intent and input entities, the DM module analyzes
them, records the current dialogue state, and then decides the direction
of the dialogue according to the contextual dialogue states.  For
example, if an entity needed to answer a question is missing, the
chatbot will ask the user to provide more information. DM also needs to
record some information (e.g., user preferences, dialogue background,
etc.) and use certain logical reasoning to give appropriate responses.
Although DM and NLU are independent modules, they affect each other's
performance in many ways \cite{harms2018approaches}. 

{\bf Knowledge Base.} After clarifying user's intention and obtaining
the necessary information, DM accesses the knowledge base to obtain the
desired answer. 
The source of the data in the knowledge base can be the
internal knowledge stored in the chatbot or the external knowledge
available through the Internet. Data is stored in a graph-structured
format in the knowledge base, where nodes are the entities and edges
are the relations. The design of knowledge bases facilitates fast,
accurate, and reliable reasoning to help DM locate the correct answers
efficiently.
Several graph machine learning algorithms, such as multi-hop reasoning
\cite{xiong-etal-2017-deeppath, abboud2020boxe, ge-etal-2023-compounding}
and graph neural networks \cite{kipf2017semi, vashishth2020compositionbased}, 
can be adopted and improve the chatbot performance
even further.

\subsection{Categories of Chatbots Based on Development Methodology}

There are three main categories of approaches to develop chatbots:
rule-based, retrieval-based, and LLM-based \cite{hussain2019survey,
thorat2020review}.  Their main differences are summarized in Table
\ref{tab:Chatbot_cat}, which will be elaborated below. 

\begin{table}[htb]
\centering
\caption{Three chatbot categories based on their development methodology.}
\label{tab:Chatbot_cat}
\renewcommand\arraystretch{1.25}
\begin{tabular}{|c|c|c|c|} \hline
\textbf{Chatbots} & \textbf{NLU} & \textbf{NLG} & \textbf{Bias} \\ \hline
Rule-Based & Pattern Matching & Predefined & Low \\ \hline
Retrieval-Based & ML Algorithm & Predefined & Low \\ \hline
LLM-Based & ML Algorithm & ML Algorithm & High \\ \hline
\end{tabular}
\end{table}

{\bf Rule-based.} Rule-based chatbots look for keywords in user input
and respond using predefined rules. They are adopted by simple
question-and-answer systems. Early chatbots were mainly rule-based, such
as ELIZA and AIML-based chatbots \cite{satu2015review}. In developing
rule-based chatbots, the design team needs to define rules manually,
which is a tedious job.  Since the content that humans can define is
limited, input queries that can be effectively replied to are limited. In
faced of situations where keywords cannot be matched, rule-based
chatbots can only change the topic using predefined sentences.  The
decision-making process of rule-based chatbots is clear, and their
responses are controllable.  Rule-based chatbots have a poor understanding
of context and language. Their answers lack novelty and could be highly
repetitive. Since keyword matching and response content are all set in
advance, the bias primarily comes from the development team. The bias
level is relatively low. 
        
{\bf Retrieval-based.} Like rule-based chatbots, retrieval-based
chatbots only give predefined answers so their answers could be
repetitive. On the other hand, such chatbots have learning capabilities.
That is, they use machine learning methods to train part of the question
understanding system.  Thus, they can choose more appropriate answers
from existing ones. Besides biases from the development team,
retrieval-based chatbots are subject to biases arising from machine
learning.  However, since their responses are predefined, the bias
problem is more manageable by humans. 
        
{\bf LLM-based.} Unlike the previous two types of chatbots, LLM-based
chatbots can generate new responses using large language models. They
use ML algorithms to understand user input and generative AI (GAI)
algorithms \cite{wang2023overview} to generate responses with a certain
degree of randomness. For example, ChatGPT is an open-domain chatbot that
can answer users' questions in different fields. They face several
challenges at the same time. First, training such chatbots requires a
lot of data and computing resources, which is costly. Second, the
mainstream LLM-based chatbots use unpredictable black-box models. They
are uninterpretable and without a logical reasoning process.
Consequently, their responses are unpredictable and difficult to
control.  They often contain inappropriate or fake content with biased
and offensive language, etc. 

\section{Bias Issues in Chatbots}\label{sec: Bias}

\begin{figure}[t]
\centering
\includegraphics[width=0.8\textwidth]{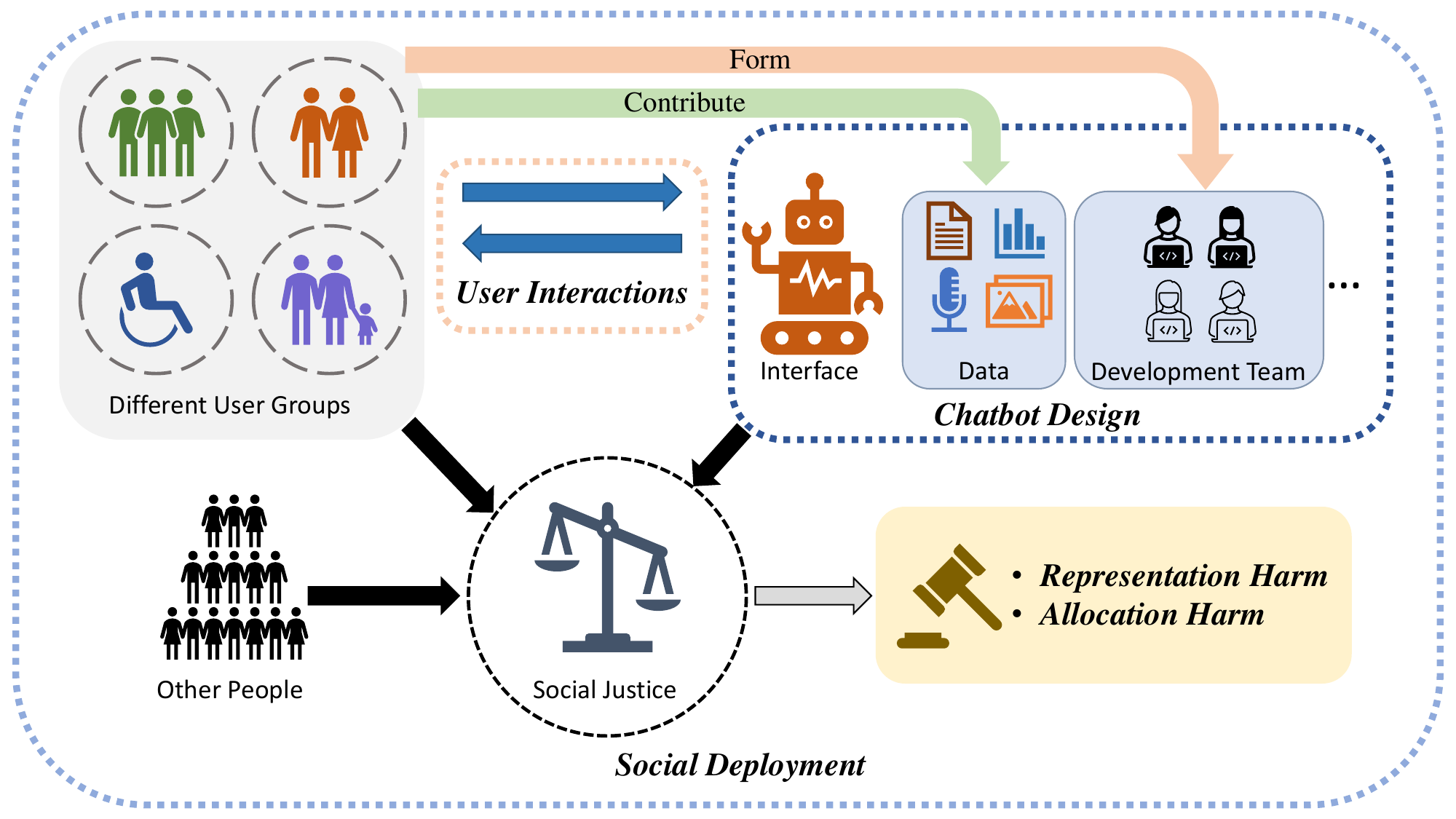}
\caption{Three bias sources in a chatbot system.}\label{fig:bias_chatbot_app}
\end{figure}

The deployed chatbots contain biases from various sources
\cite{beattie2022measuring}.  We categorize them into
three types for ease of analysis. As depicted in
Fig.~\ref{fig:bias_chatbot_app}, they can arise from: 1) chatbot design,
2) user interactions, and 3) social deployment.  First, a chatbot
development team is made up of people from different educational and
cultural backgrounds.  Their personal biases will have an impact on the
designed chatbot system. A chatbot system is composed of an external
user interface and several internal modules. The user interface design
is directly influenced by the development team.  The internal modules
rely on training data and ML algorithms.  Different data source groups
contribute to data acquisition and annotation, which can be biased as
well. The final data used for training needs to be screened by the
development team, leading to another bias source.  Second, after a
chatbot is deployed, it interacts with users and biases can be enhanced
in the interaction process. The bias can even affect user's view and
value. In addition, users may become part of the development team and
contribute to data annotation in the future, which makes bias generation
a vicious circle.  Third, biases may come from the environment where
chatbots are deployed. For example, people's attitudes toward chatbots and
the way chatbots are used can lead to biases.  A biased chatbot
system and people affected by the bias may result in representation and
allocation harms to social justice. These topics are the main focus of
this paper. They will be detailed below. 

\subsection{Biases from Chatbot Design}

\begin{figure}[t]
\centering
\includegraphics[width=0.8\textwidth]{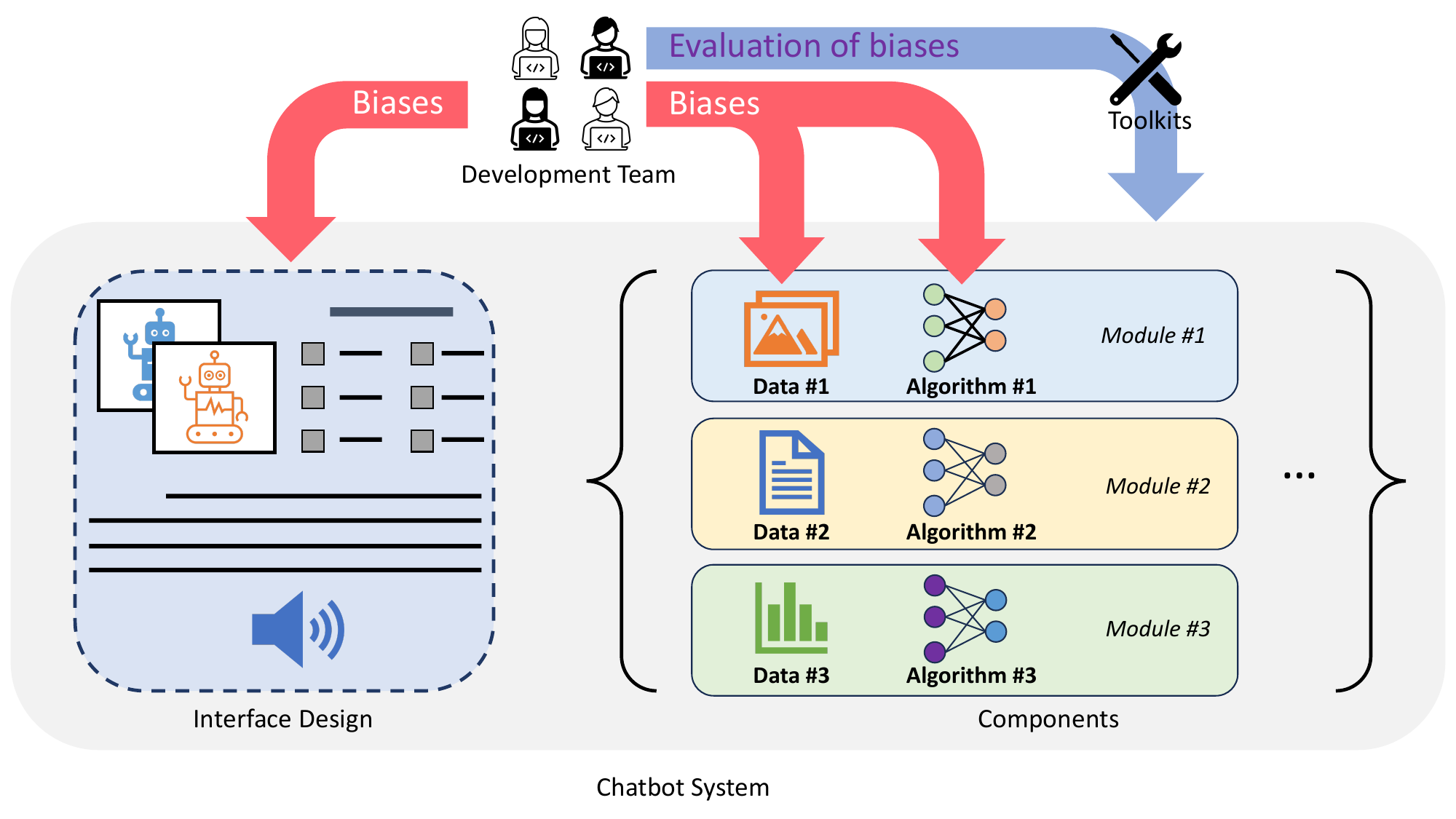}
\caption{Biases in the design of chatbot systems.}\label{fig:bias_chatbot_design}
\end{figure}

Fig.~\ref{fig:bias_chatbot_design} gives an overview of biases in
designing chatbot systems. A chatbot is composed of a user interface
module and several internal components. Each of them can have several
bias sources.  The development team may pass its biases to each
component. On the other hand, they can utilize some toolkits to mitigate
biases. 
        
\subsubsection{Biases in Development Team} 

People are biased \cite{cain2008everyone}, and developers are no
exception. Individual biases may be influenced by personal experience,
family upbringing, culture, education, etc. Developers' cognitive biases
can affect developed software \cite{mohanani2018cognitive}. Our brains
tend to simplify the world for decision efficiency, which results in
cognitive biases \cite{korteling2018neural}. Being subject to cognitive
biases, developers may ignore certain factors and/or situations in
designing rules or selecting training data and/or input features.
Another example is that an all-male development team often sets the
chatbot gender to a female. Some of the biases are explicit, while
others could be unconscious and implicit. For user groups with similar
biases as developers, there is an attraction effect between them and the
development team \cite{zabel2021bias}. They may have a better user
experience with the chatbot.  On the other hand, implicit or unconscious
biases may worsen the user experience of other groups. 
            
One way to mitigate such biases is to increase the diversity of the
development team \cite{meissner2017effect, feine2020gender}.  Although
it would be ideal to include subgroups among users to reduce the
marginalization of minorities, this is however too difficult to
implement. In practice, the development team may hire people of
diversified backgrounds, with different ways of thinking and
perspectives, and with different expertise to control biases.  Besides,
the development team can use some toolkits to evaluate biases of the
system and take steps to mitigate them. For example, FairPy
\cite{viswanath2023fairpy} is a toolkit for quantitative bias evaluation
in pre-trained LLMs like BERT. It takes different types of biases into
account, such as race, gender, age, etc.  AI Fairness 360
\cite{bellamy2018ai} is another toolkit aiming at transitioning fairness
research to industrial settings and providing a general framework in
fairness algorithm sharing and evaluation. It helps developers detect
and mitigate biases in ML models.  As another toolkit, Aequitas
\cite{saleiro2018aequitas} takes application scenarios and non-technical
people into account.  It sets up several bias and fairness metrics
associated with multiple population subgroups in the ML workflow.  As a
scalable toolkit, LiFT \cite{vasudevan2020lift} can incorporate the bias
measure and mitigation mechanism in ML systems operating on distributed
datasets. 
        
\subsubsection{Biases in Interface Design} 

Unlike general ML systems, chatbot systems have unique interfaces.  To
make chatbots more human-like and improve user experience, companies add
some attributes to chatbots, such as names, resumes, descriptions,
avatars, and voices.  Most users do not understand the
architecture of chatbots and can only interact with chatbots through
these attributes.  Then, the interface design has an impact on user
perception. However, a chatbot's attributes can contain stereotypes and
biases, with the gender bias being the most prominent
\cite{costa2018conversing}. 
            
A UNESCO report \cite{west2019d} pointed out that the vast majority of
voice assistant chatbots (e.g., Siri, Alexa, etc.) are designed to be
female. They have feminine names with female voices and appearances as
the default. A study \cite{feine2020gender} examined 1,375 chatbots on
the chatbots.org website and found that most chatbots were designed as a
female.  This is especially true in the sector of customer service and
sales.  In real life, many workers in the service industry are women
\cite{rho2020basic}, and people often have stereotypes of women as
helpful, gentle, accommodating, and nurturing.  When chatbots appear in
people's lives as service providers, developers set them as women by
default to increase affinity. 

On the other hand, some robots, such as those designed for
stereotypically male tasks that need to show strength and leadership,
use male personas \cite{costa2019ai}.  Due to people's gender
stereotypes, names, voices, body shapes, and facial cues of robots may
affect user's impression and trust in robots \cite{nass1997machines,
eyssel2012s, bernotat2021fe, bernotat2017shape}. They may have an impact
on customer satisfaction \cite{seo2022female}. Such a setting may be in
company's business interest, since users prefer chatbots that present
stereotypes of specific roles \cite{mcdonnell2019chatbots}.  Another
reason for chatbot feminization could be the low proportion of women in
the chatbot development team \cite{folstad2017chatbots}.  Apparently,
designing a chatbot to be a female is more appealing to an almost
all-male development team. 
            
However, the design, which is beneficial to business, could be harmful
to our society. The ubiquity of chatbots designed according to
stereotypes will reinforce people's stereotypes. For example, in the
case of voice assistants, the feminization of voice assistants
contributes to stereotypes about women. As chatbots, they will try their
best to meet the needs of users. However, female chatbots are more
likely to be targets of sexual harassment and abuse
\cite{brahnam2012gender, woods2018asking}. When faced with inappropriate
requests such as sexual harassment and bullying, most of them choose to
avoid or pretend not to understand \cite{curry2018metoo}. These
responses appear to set an example for women to accept abuses and teach
them how to respond to unjustified demands. This kind of responses turns
the sadistic behavior into an acceptable behavior and reinforces the
stereotype that women are accommodating and submissive. The consequence
clearly brings harms to our society.  Nowadays, some companies are aware
of this issue and have taken actions to control the gender bias in
chatbots, such as providing male voices instead of the default female
voice and making chatbots appear tough in the face of inappropriate
language. However, the interface design is still a source of biases in
chatbot systems. It can contain many forms of biases, of which the
gender bias is an obvious one. 

\subsubsection{Biases in Internal Components} 

A chatbot system consists of many internal components such as multimodal
processors, natural language processing, dialogue management, etc. Each
is responsible for a specific task as shown in
Fig.~\ref{fig:chatbot_arc}. The design of each component can lead to
biases.  For rule-based modules, biases are mainly from limited rules
and predefined responses. They are less and easier to control.  For
ML-based components, the use of ML algorithms makes the system more
capable yet biased. ML algorithms are used in a wide range of fields
\cite{sarker2021machine}, including important and sensitive areas such
as healthcare \cite{mhasawade2021machine} and recruitment
\cite{kochling2020discriminated}, nowadays. Their bias problem has
received more attention. There are many papers on the bias and
fairness issues in AI and ML. Some analyze possible bias sources in
general AI systems \cite{roselli2019managing, nelson2019bias,
srinivasan2021biases}, and attribute them to data sources and AI/ML
algorithms. Others discuss specific types of bias in AI systems, such as
the racial bias \cite{noor2020can, kostick2022mitigating} and the gender
bias \cite{nadeem2020gender, domnich2021responsible}. There are also
papers on specific application fields such as healthcare
\cite{parikh2019addressing, cirillo2020sex} and education
\cite{baker2021algorithmic, kizilcec2022algorithmic}. Generally
speaking, the bias in the ML-based modules mainly come from data and
algorithms, e.g., data collection and labeling, feature selection, the
way data are used in ML algorithms, etc. \cite{mehrabi2021survey}.
Biases in one component can have an impact on the performance of the
other, and the contribution of each component to the overall biases in
the system is often difficult to determine. Combinations of individual
components that meet fairness metrics can also exhibit biases. To
control the biases of the whole system, it is important to mitigate the
bias in each component and evaluate the biases of the whole system. 

{\bf Biases in Data.} The bias may come from the data source or from
people's collection and labeling of data. The data source is affected by
human biases such as reporting bias, selection bias, etc.
\cite{gordon2013reporting, winship1992models}. It is recorded for
various reasons. It may neither reflect the actual distribution nor have
a proper balance among subgroups. Next, training data are selected from
the data source and annotated. In the selection process, there are
sampling bias, in-group bias, measurement bias, etc.
\cite{fischer2016group, delgado2004bias}.  Thus, the distribution of
training data may not be the same as that in the application scenario.
Furthermore, the data labels can be affected by the world views of
annotators, resulting in experimenter's bias and confirmation bias
\cite{klayman1995varieties}. 
            
{\bf Biases in ML Models/Algorithms.} Features considered by ML models
often lead to biases \cite{kordzadeh2022algorithmic}.  Even if sensitive
features are not directly used as input, some features that are highly
correlated with sensitive features will allow models to learn bias.
During training, biases in the training data are amplified.  In a
general ML system, the aggregation bias \cite{suresh2021framework} may
occur under the influence of broad categories of data groups, and models
may draw wrong conclusions about individuals due to group trends. In
addition, the evaluation of models can also lead to bias. Sometimes, the
evaluation criteria do not accurately reflect the desired goal. In a
chatbot system, almost every module can be implemented using ML
algorithms to improve performance, so they may all contain certain
biases. 

{\bf Biases in Multimodal Processors.} As AI systems become more
advanced, people are no longer satisfied with the input and output of a
single modality, and multimodal processors are used to handle the
multimodal communications between humans and robots. The combination of
multiple modalities will often increase model biases and compromise
fairness \cite{booth2021bias}.  Biases can be hidden in both algorithms
and training data for each modality. For example, the automatic speech
recognition (ASR) technique, which enables chatbots to recognize and
interpret human speech, is an essential component of voice assistants.
However, ASR systems can have biases such as gender, age, and
regional accent biases \cite{feng2021quantifying}. They may come from the
composition of the corpus, the mismatch between the pronunciation and
speech rate of users and the training data, or biased transcriptions,
etc. Another example is image captioning. The ML model converts images
to text, which can be affected by biases in the image context. For
example, men are associated with the snowboard while women are
associated with the kitchen in training images
\cite{hendricks2018women}. In chatbot systems, multimodal processors are
usually directly connected to the interface.  When converting multimodal
user input into text, their biases may distort user input or omit
important information, which affects subsequent dialogue understanding
and answering. On the other hand, when converting generated text to
multimodal output, the bias may lead to inappropriate output content and
affect user experience. 

{\bf Biases in NLP Models/Algorithms.} NLP is a branch of AI that
aims to equip computers with the ability to understand and generate
human languages.  There are quite a few recent papers on the bias and
fairness issues in NLP \cite{garrido2021survey, hovy2021five,
chang2019bias, blodgett2020language, dixon2018measuring,
elsafoury2023bias}.  They show that NLP models can contain biases and
there are methods to mitigate them. For example, word embedding is an
important technique that represents words in vector form to facilitate
computer understanding of human language. However, word embedding models
often contain human-like biases \cite{abbasi2019fairness}, such as
associating men with computer programmers while associating women with
homemakers \cite{bolukbasi2016man}. The use of word embedding in NLP
downstream tasks may amplify certain biases.  The cosine
similarity-based Word Embedding Association Test (WEAT)
\cite{caliskan2017semantics} and its variants \cite{lauscher2020general,
dev2020oscar, guo2021detecting} have been proposed to measure and
mitigate these biases.  They are applied to models such as Word2Vec
\cite{mikolov2013efficient} and GLOVE \cite{pennington2014glove}. 

Besides word embedding, there are other biases in NLU and NLG tasks. For
the NLU task, biases in coreference resolution have received much
attention \cite{rudinger2018gender}. Methods like debiased word
embedding and data augmentation can mitigate these biases effectively 
without affecting performance much \cite{zhao2018gender,
park2018reducing}. For NLG tasks, biases can come from deploying systems
and decoding techniques \cite{sheng2021societal}. Methods for
measuring and controlling such biases have been proposed as well
\cite{sheng2020towards, dhamala2021bold, parrish2021bbq}. 

With the advancement of NLP technology, large-scale language models are
trained on a wide range of corpora to master general human languages.
Although these pre-trained models can be used as starting points for
downstream NLP tasks to improve efficiency, they can be biased
\cite{schramowski2022large}. For example, as a large-scale natural
dataset in English, StereoSet \cite{nadeem2020stereoset} evaluates
biases in pre-trained models and demonstrates that most mainstream
pre-trained models exhibit strong stereotypical biases. These biases are
propagated to downstream tasks, affecting the performance of downstream
models. 

\subsection{Biases from User Interactions}

\begin{figure}[t]
\centering
\includegraphics[width=0.8\textwidth]{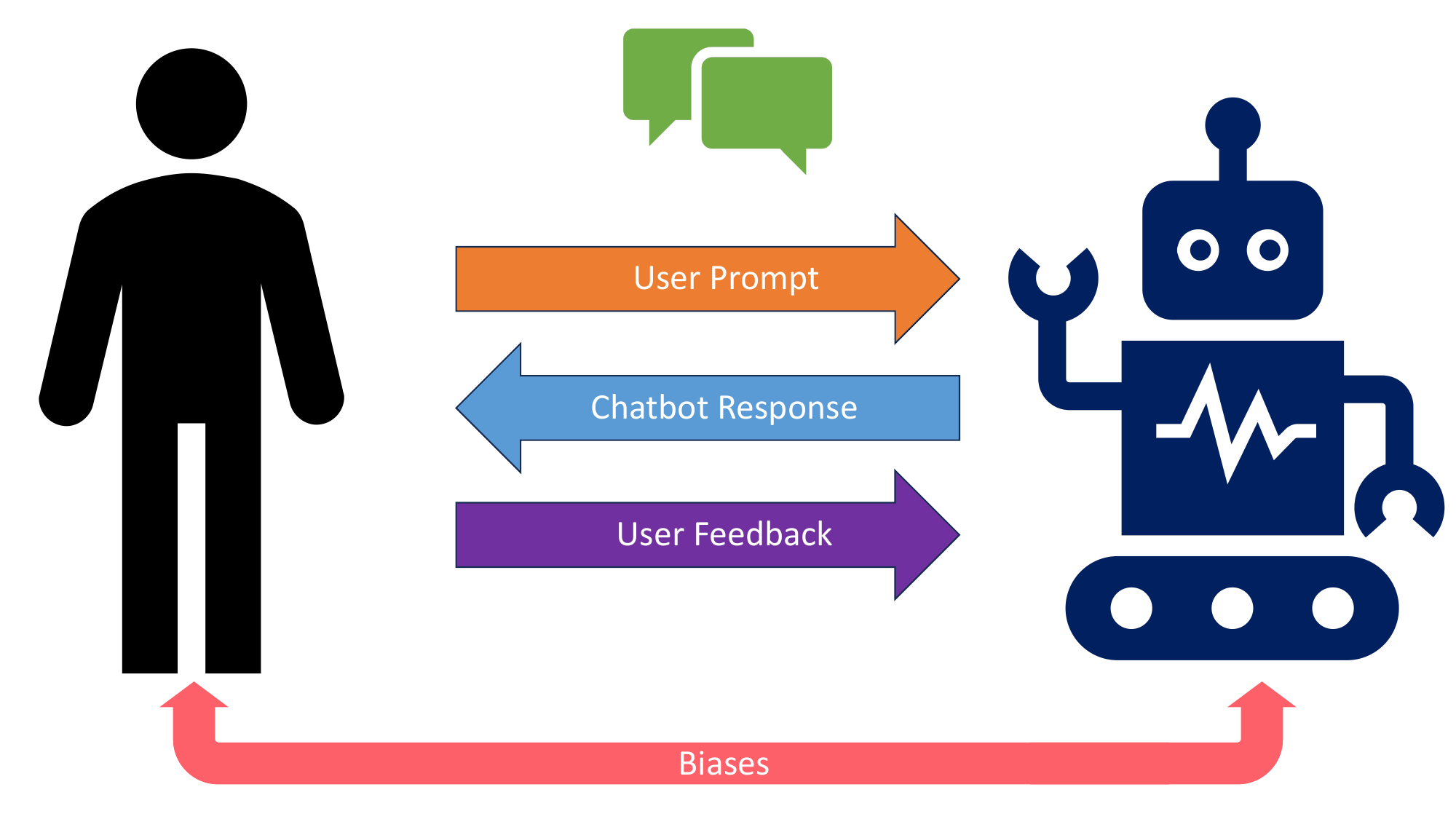}
\caption{The bias coming from the interaction of users and a chatbot.}
\label{fig:bias_user_interaction}
\end{figure}

In the chatbot development phase, biases mainly originate from
developers, training data, and algorithms. When the service is launched,
the chatbot interacts with users. It gets prompts and feedback from
users. It learns from interactions, which makes it more capable but
introduces bias. This is a significant difference between chatbot
systems and other traditional ML systems.  The interactions between a
chatbot and users are illustrated in
Fig.~\ref{fig:bias_user_interaction}. Users first give the chatbot a
prompt to start a conversation. The chatbot will return with a response.
Users can grade the response, and ask for regeneration or give a new
prompt. In this way, users and the chatbot can exchange information
with their biases being propagated mutually. This topic is elaborated
below.
        
{\bf Biases in User Prompts.} Unlike some ML systems where users accept
predictions unilaterally, chatbots respond based on prompts from the
user. Users' prompts can be in multiple languages. Automatic
determination of the language is an important step for further
processing. Language identification (LID) is usually used to
detect the input language type \cite{mcnamee2005language}. An ideal LID
tool should be unbiased to any language in terms of inference time,
response content \cite{ambikairajah2011language}, etc. As pointed out in
\cite{ball2021differential}, dialects can cause biases in NLP tasks. For
prompts that contain informal language, such as dialects, chatbots may
misunderstand and respond with biases.  Also, keywords in prompts are
obvious. For chatbots that learn stereotypes from massive data, certain
keywords may trigger stereotypes in the model and make the model
generate stereotyped results.  For example, a prompt containing the keyword
``Muslim'' might yield results related to ``Terrorist''
\cite{abid2021persistent}. Sometimes, although there are no obvious
keywords in the prompt, specific chat topics can lead to bias
\cite{si2022so}. For example, women are more likely to appear in the
topic of ``family'' in GPT-3 \cite{lucy2021gender}. Also, input of
similar meaning in different prompts may lead to different results.  A
prompt could be biased, and a chatbot that cannot perceive the bias may
respond to biases given by the prompter.  Even the prompt is not biased,
the chatbot may show certain tendency. To get fairer results, chatbots
should be able to recognize biases in user prompts.  Users may need to
think about chatbots' derivation process instead of accepting the
responses blindly. 
        
{\bf Biases in User Conversation.} Learning from conversations with users
is an important way for some chatbots to expand their knowledge base and
become more powerful.  On the positive side, this can enhance their
understanding of human language and allow them to generate more accurate
responses. However, the user group is large and complex, and user
prompts can contain biases and/or stereotypes.  Some users may even
deliberately guide chatbots to express biased speech. If the chatbot
learns in a wrong way, it could be harmful. Microsoft's chatbot Tay was
an example. The rapid change of its conversational style demonstrated
the destructiveness of continuous learning from biased user
conversations. Thus, it is important to check the responses generated by
the chatbot with a systematic approach at a certain frequency after its
initial deployment. 
        
{\bf Biases in User Feedback.} Many systems allow users to give some
feedback on the response, such as asking to regenerate the response or
rating the response after a chatbot responds.  This can improve the user
experience. The feedback can help chatbots understand what kind of
results are more accurate and more in line with users' expectations to
enable the machine to generate more satisfying or personalized responses
for users in the future. When chatbots generate biased, fabricated, or
incorrect responses, negative feedback from users can help them correct
mistakes. Without feedback, chatbots may not be able to detect and
correct their own biases timely, and these biases may be further
amplified in subsequent learning.  However, users have various biases
and people favor responses that cater to their biases
\cite{mcdonnell2019chatbots}. As a result, they tend to give higher
scores to responses that contain biases. When chatbots use biased user
feedback to learn and improve responses, future responses will have
increasingly severe biases. Thus, how to use user feedback to alleviate
biases in chatbot systems is an important topic. 

{\bf Biases in Chatbot Responses.} Biases are passed on to users in
chatbots' responses. Generated responses of some chatbots are always
confident and smooth, regardless of whether they are correct and/or may
contain inappropriate information \cite{taecharungroj2023can}.  This
makes it difficult for users to distinguish authentic information from falsified
information. Young children, in particular, lack the ability to
recognize biases in chatbot responses and to judge truth from fiction. It thus poses
a challenge to chatbot's educational applications
\cite{kasneci2023chatgpt, kooli2023chatbots}. One solution is to ask the
chatbot to provide references and the reasoning process along with their
responses. The additional information helps users judge whether the
content is reasonable and correct. However, some modern chatbots, such as
ChatGPT cannot provide proper references \cite{gravel2023learning}. For
example, when users ask ChatGPT to give reference links, most of them
are actually irrelevant \cite{day2023preliminary}. The opacity makes it
difficult for users to judge the accuracy of received responses.
        
{\bf Presentation and Ranking Biases.} Chatbots play a role similar to
search engines that may have biases in delivering information to users
\cite{baeza2018bias}.  Unlike traditional search engines, they extract
and summarize the information on the Internet (rather than listing raw
URLs). Since there is too much information on the Internet, it is
difficult for chatbots to present all of them.  Which information to
present and whether the presented information is balanced are determined
by the algorithms.  The information not presented cannot be received by
users. All of these lead to presentation bias. The presented information
by chatbots may be ranked or with a certain focus, causing ranking bias. 

\begin{figure}[t]
\centering
\includegraphics[width=0.8\textwidth]{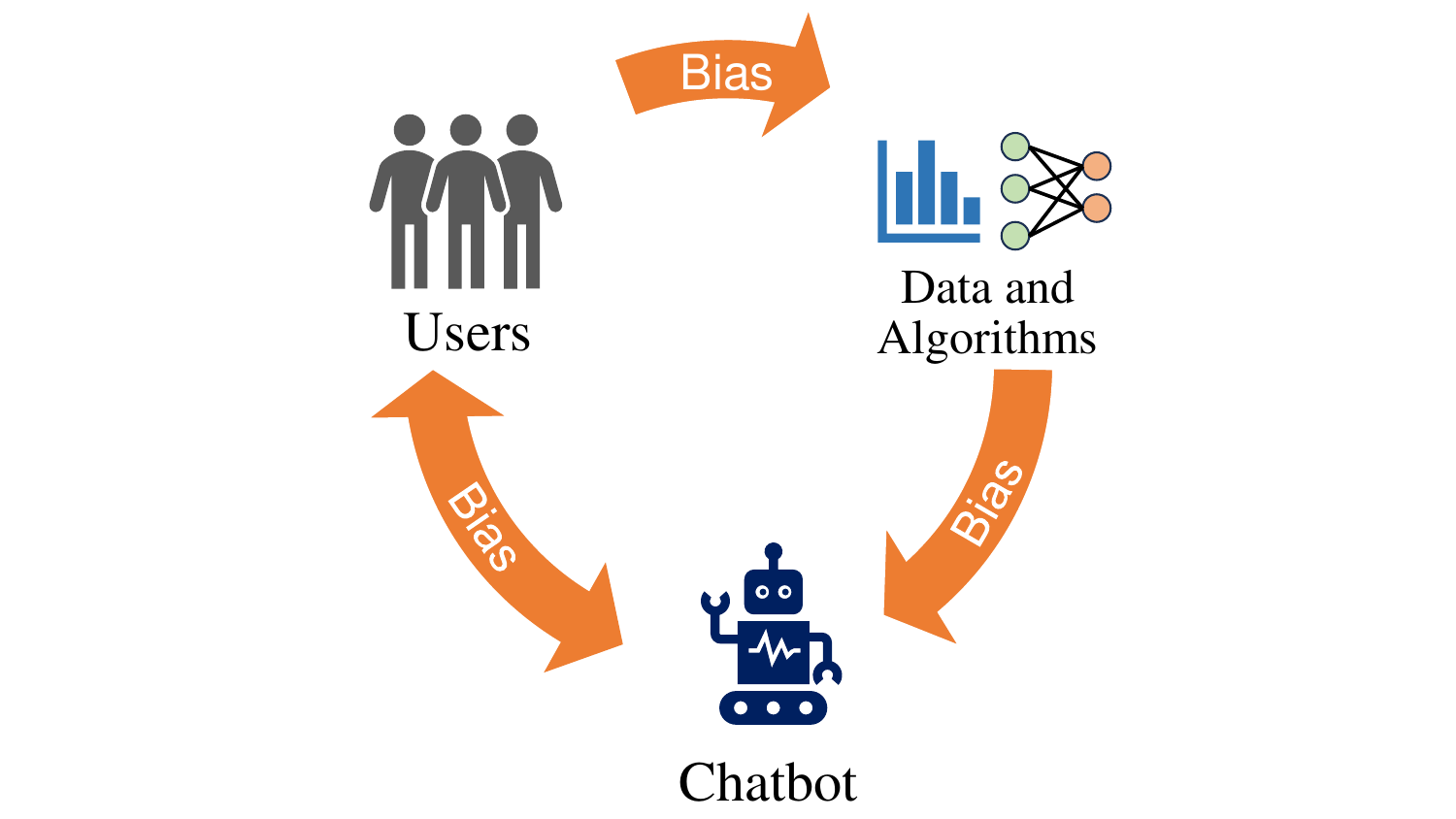}
\caption{Illustration of the vicious bias circle in chatbots.} 
\label{fig:vicious_circle}
\end{figure}

{\bf Vicious Bias Circle.} When people have long-term conversations with
biased chatbots, the passed biases can affect their worldviews. This is
especially severe for children. The biased worldviews will 
affect data collection and annotation, model training, and chatbot
development.  In this way, biases will become more serious, forming a
vicious circle as shown in Fig.~\ref{fig:vicious_circle}. 

\subsection{Biases from Social Deployment}

Technologists try to mitigate biases in the product design to make the
chatbot fairer. However, failure to understand the interaction between
the chatbot and its social environment may result in the contextual
mismatch \cite{selbst2019fairness}. The chatbot is deployed in a
community as shown in Fig.~\ref{fig:bias_chatbot_app}. There are
multiple user groups interacting with the chatbot. Some of them may
never use the chatbot before.  Different chatbots have different social
contexts, and their fairness consideration cannot be separated from the
social context.  Everybody in the community has the potential to
influence the chatbot and, in turn, to be influenced by people who
interact with the chatbot.  Chatbots that meet the fairness criterion
before deployment may become biased in a particular community later
since biases may stem from people's attitudes to chatbots, different
user group compositions, different chatbot usages, etc.  Thus, it is
essential to consider the chatbot deployment environment to mitigate 
the bias. We will focus on biases arising from people's attitudes,
application background, and solution selection below.
        
{\bf Biases in People's Attitudes.} When a new technology is applied in a
social environment, human perception of the technology may lead to
biases that have a great impact on human-machine cooperation. Even if
a technology meets fairness metrics, whether it will be biased in
practice depends on how people use it. The same is true for chatbots.
User's attitudes toward a chatbot determine how they use it and the
impact it can have.  Since chatbots reason differently from humans and
make different mistakes, they would perform better when humans and
chatbots work together. On the other hand, a chatbot's speech may unduly
influence human behavior, and the way people use chatbots may go
beyond the expectations of their designers.

Some people may unconsciously believe in chatbots for various reasons,
such as not being confident enough about themselves, over-believing in
the answers of chatbots, fear of taking responsibility
\cite{wang2023interactive}, etc. All of them lead to the automation
bias, which allows chatbots to propagate wrong knowledge or fake content
more easily. Some of today's chatbots speak in a very confident tone,
regardless of whether the content is correct or not, which exacerbates this
phenomenon. 

The automation bias is common in AI systems. For example, results of the
Allegheny Family Screening Tool (AFST), a predictive model for child
abuse, may cause staff members to question their own judgment
\cite{eubanks2018automating}. The Correctional Offender Management
Profiling for Alternative Sanctions (COMPAS), which predicts the risk of
recidivism, is another example. Its designers did not intend for the
model to determine a person's prison time. Yet it was used by judges in
sentencing \cite{angwin2022machine}. People tend to rely too much on AI
systems because they may subconsciously think that robots without
emotions would give unbiased conclusions. They doubt themselves and
follow chatbot's opinions when they have different viewpoints.  Similar
issues arise in chatbot systems that are used to predict or provide
decision-making advice. 

The chatbot output is in human language. It can be used to write
articles. If an author relies too much on chatbots, the resulting
article may contain fake or plagiarized content. The opaqueness of
sources and the flamboyance of the article make it difficult for authors
to spot the errors. Once caught, the author may attribute the error to
the chatbot.  However, chatbots should not be held accountable for the
mistake. It is important to have an appropriate accountability mechanism
in place \cite{van2023chatgpt}.  
        
Besides, people may have an aversion to chatbots. This happens for a
number of reasons. For example, some might be happy working with a
chatbot at first. However, after the chatbot made severe mistakes, they
did not want to continue working with any chatbot. Others may be
influenced by media propaganda to form stereotypes about chatbots.
Furthermore, some do not trust third-party chatbot companies and cannot
accept their domain being influenced by them.  Others worry that
chatbots will take their jobs \cite{haque2022think} and try to exclude
them as much as possible.  Without the aid of a fair chatbot, people may
struggle to detect their own implicit biases. Whether believing in the
chatbot system too much or resisting it too much is not helpful to
building a cooperative relation between humans and chatbots.
Understanding and analyzing how people perceive chatbots in social
environments are needed to control potential biases. 
        
\begin{figure}[t]
\centering
\includegraphics[width=0.8\textwidth]{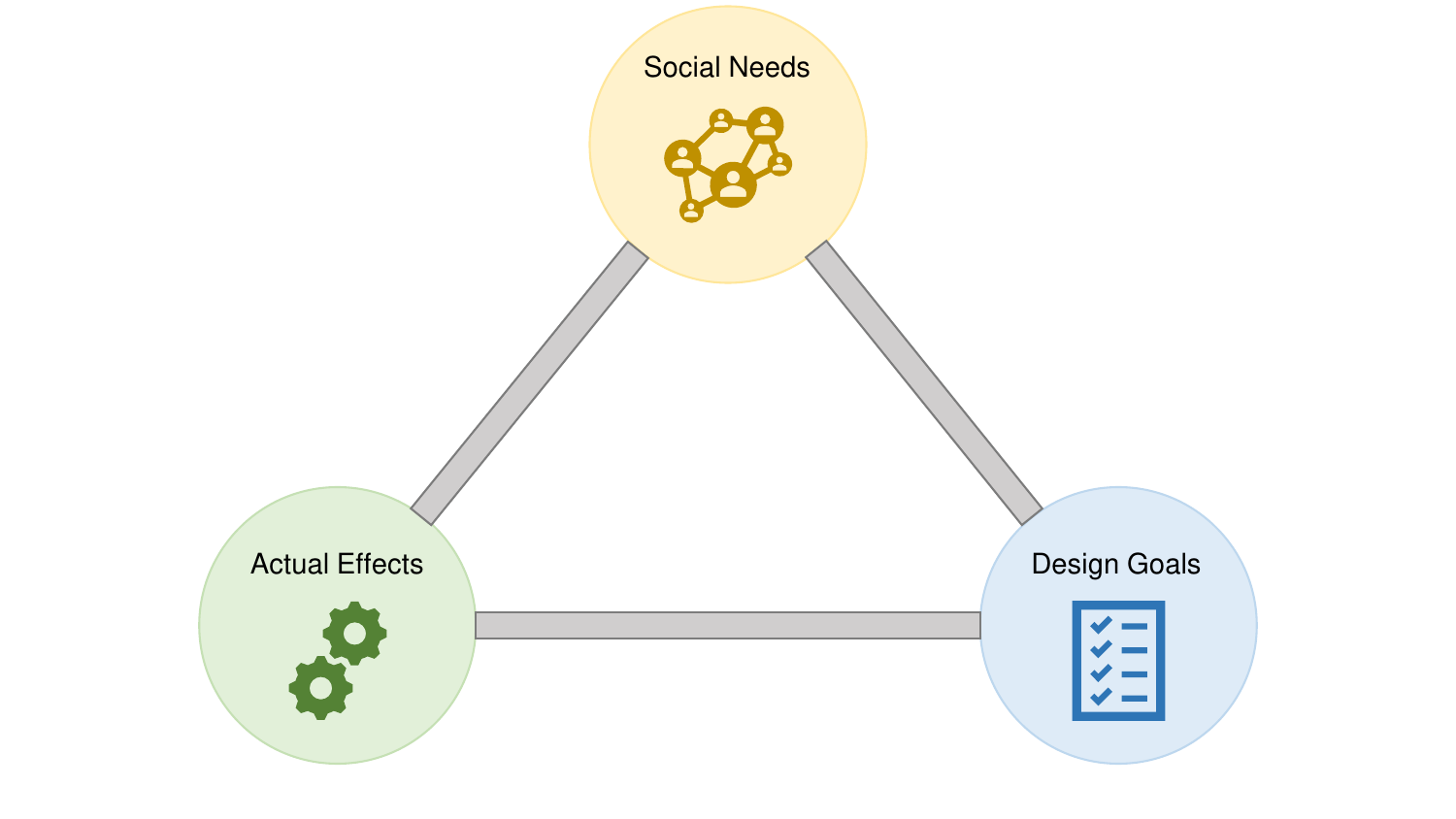}
\caption{Three key factors of deploying chatbots in a specific 
application domain.}\label{fig:bias_app_background}
\end{figure}

{\bf Biases in Application Domain.} Different chatbots are designed
for various applications \cite{janssen2020virtual}. Open-domain chatbots
can talk to people without being limited by topics and domains.
Domain-specific chatbots master the knowledge in specific domains, and
they are designed for specific tasks.  Their social environments and
served user groups can be quite different.  Three key factors of
deploying chatbots in a specific application are shown in
Fig.~\ref{fig:bias_app_background}. They are social needs, design goals,
and actual effects. Before designing a chatbot, the development team
should understand social needs and establish appropriate design goals
accordingly. Then, the chatbot will be designed based on the design
goals. The final effects of the chatbot product should meet design goals
and social needs. The three factors should be consistent to minimize the
biases in a specific social environment.  Design goals play an important
role as a mediator between actual effects and social needs.

To deploy a chatbot in a social group, understanding social needs is the
first priority. The development team should set up the design goals and
choose proper fairness metrics according to the needs so as to design a
chatbot that can be well integrated into the society. However, in
reality, programmers tend to emphasize portability. Some models may be
shared under different social needs, such as many pre-trained LLMs in
NLP. Sometimes, models may be designed for a specific user group, and
their design goals are based on some assumptions about the social
context. When transplanted to the other user group, these assumptions
may not hold and there is inconsistency between design goals and social
needs. Then, biases could appear. For example, a chatbot is designed for
users in a country that has specific training data and design goals.
If the same chatbot is deployed in a more conservative country, its
responses may be considered offensive and not in line with user
expectations. Clearly, the conflict between social needs and design goals
leads to biases. 

Inconsistency between design goals and actual effects can lead to
the bias. The development team needs to model and implement a chatbot
according to design goals, including data collection, feature selection,
rule-making, social equity modeling, etc. Improper modeling may result
in poor models. For example, a development team wants to design a
companion chatbot that understands and responds to user emotions.  The
design may allow the chatbot to understand users' emotions through
real-time facial emotion recognition. However, facial expressions may
have different emotional implications in various cultures. The use of
facial expressions alone to judge emotions is problematic in practical
applications \cite{chen2018distinct}.

{\bf Biases in Solution Selection.} When people get used to a solution, it
is often difficult to think of new ways to solve the problem. For a new
problem, development teams will naturally give priority to the solutions
that they are more comfortable with. This bias may come from propaganda
in the society or people's experience. For example, the recent
popularity of ChatGPT has motivated people to apply it to various
fields. LLM-based chatbots have become mainstream solutions to many
problems \cite{aydin2022openai, haleem2022era, wang2023chatcad}.
However, they may not be optimal for some tasks. For example,
traditional chatbots with predefined output results may be more economical
for customer support. They save resources and have a lower risk in the
bias.  While results generated by generic models such as ChatGPT are
diverse, they may not be accurate and correct. For chatbots in the
medical field, although LLMs such as ChatGPT are applicable, their
opaque reasoning process can be challenged.  Besides, there is a risk in
the leakage of personal sensitive data during the dialogue process between
humans and chatbots. Traditional methods could be more robust and
privacy-preserving.  While advanced methods marginalize some user
groups, traditional methods can be tailored to them and offer a better user
experience. 

\subsection{Harms from Negative Biases}

Biases may not always be bad, but negative biases can result in serious
harms to our society. Harms caused by biases can be divided into two
types: the allocation harm and the representation harm
\cite{crawford2017trouble}.  The allocation harm occurs when the machine
is used to make decisions and allocate resources, which significantly
benefit some groups against others.  The representation harm occurs when
the machine reinforces associations or stereotypes between certain
representative traits and some people groups.  The differences between
the two are summarized in Table \ref{tab:Chatbot_harm}. 

\begin{table}
\centering
\caption{Comparison of the allocation and the representation harms.}\label{tab:Chatbot_harm}
\renewcommand\arraystretch{1.25}
\begin{tabular}{|c|c|} \hline
\textbf{Allocation Harm} & \textbf{Representation Harm} \\ \hline
Short-term & Long-term \\ \hline
Limited range & Wide range \\ \hline
Evident & Elusive \\ \hline
Readily quantifiable & Challenging to quantify \\ \hline
\end{tabular}
\end{table}

{\bf Allocation Harm.} The allocation harm often has a specific
application scenario and happens quickly. It happens when decisions
made by people are influenced by biased machines. As chatbots become
smarter, people may rely on them to analyze problems. If chatbot's
analysis is biased and influences human decisions, the allocation
harm arises. For example, existing research suggests that using machines
to guide health decision-making may result in the allocation harm to
African-Americans \cite{rajkomar2018ensuring, obermeyer2019dissecting}.
With the popularity of LLM-based chatbots, we see a trend to use
chatbots for automatic diagnosis. When a biased chatbot is used for
diagnosis, different age or ethnic subgroups with the same symptom may
have different diagnostic recommendations. This affects people's judgment
on the condition and subsequent medical decision-making, leading to the
unfair distribution of medical resources. The allocation harm leads to
the short-term unfair distribution of social resources and affects
a limited range of people. It is easier to detect and quantify. 

{\bf Representation Harm.} The human society has many stereotypes that
associate representative characteristics with specific groups. In the
context of chatbot systems, users communicate frequently with chatbots.
If some characteristics are always associated with specific groups in
the conversation with the chatbot, users' worldviews can be subtly
changed and stereotypes will be exacerbated. The wider the user base,
the more far-reaching this effect will be. The representation harm also
occurs in chatbots when affected users participate in developing
chatbots, such as data recording and annotation, model design, and
system evaluation. The representation harm takes a longer time to develop
and lasts for a longer time. It is elusive and relatively difficult to
detect, change, quantify, and track. Although the representation harm is
less obvious, it cannot be underestimated. It may change people's
perception of the world and the future direction of our society. 

\subsection{Bias Mitigation in Chatbots}

As people pay more attention to biases in AI systems, various
methods to mitigate biases have been proposed. They are generally
categorized into pre-processing, in-processing, and post-processing of
ML algorithms. Similar methods can also be effectively applied to
address biases in chatbot systems. For chatbot systems, biases can be
mitigated in three stages: 1) preparation, 2) development, and 3)
optimization.  In this section, we will explain how biases can be
mitigated in each stage of chatbot systems.

{\bf Preparation.} To design a chatbot system for a specific
application, background research in the preparation stage is important.
Understanding the context of the application helps identify the social
needs, set reasonable design goals, and assemble a balanced development
team. It mitigates the biases caused by contextual mismatch.
Preprocessing is important in the preparation stage to mitigate
underlying biases in the training data by aligning the statistical
distribution of the datasets with the real-world application scenarios.
For example, several data preprocessing techniques, such as
oversampling, undersampling, data augmentation, etc., can be used to
mitigate the bias in the training data \cite{qraitem2023bias}.  Removing
sensitive features and relabeling some samples in the dataset before
training also help mitigate biases \cite{linardatos2020explainable}. 

{\bf Development.} During chatbot development, developers need
to design a proper user interface and algorithms for different internal
components.  For the interface, suggestions and feedback from experts
and volunteers in different user groups are important for designing and
evaluating.  To mitigate biases from internal components, adopting
models that have better performance on specific fairness metrics is
beneficial.  In addition, having an interpretable and transparent
decision-making process is crucial to bias alleviation.  After choosing
certain models, techniques are available to further mitigate biases.
For example, to mitigate the gender bias in NLP tasks, learning
gender-neutral word embeddings, using constrained conditional models,
and utilizing adversarial learning are helpful \cite{sun2019mitigating}.
More generally, adopting fairness-aware classifiers
\cite{zafar2017fairness}, adding regularization and constraints, and
ensemble different models are useful in mitigating biases in ML
algorithms. In the development phase, the fairness toolkits mentioned in
the previous section are useful for developers to evaluate their
algorithms and make adjustments.

{\bf Optimization.} As an AI system that primarily interacts
with humans, chatbots require further optimization and maintenance after
deployment \cite{mokander2023auditing}. When communicating with users,
biases that were not considered may occur \cite{solaiman2023gradient}.
Developers need to optimize the system and mitigate such biases to
prevent further harm. For example, they can use a rule-based model or
train a new module to detect biased prompts from users. Once the model
finds a biased prompt, the chatbot can correct the prompt or deny the
request. When learning from human interactions, human supervision or
bias detection models can be introduced to filter out biased information
so the chatbots will not learn from the misinformation. Besides, after a
system is deployed, it is important for developers to explain the proper
usage of the chatbot and inform users of potential risks and the scope
and capability of the chatbot.  Developers may also check the misuse
problems to avoid biases regularly.

\section{Fairness in Chatbot Applications}\label{sec: Fairness}

With the advancement of AI and the occurrence of many unfair cases
caused by AI applications with bias and discrimination
\cite{ferrara2023fairness}, people have paid more attention to the
fairness of AI in real-world applications \cite{caton2020fairness}. As a
specific AI system that interacts with humans, fairness issues arise in
chatbot applications \cite{li2023fairness, zhang2023chatgpt}.  A fair
system means one without negative bias and discrimination. An
unfair chatbot system may produce biased or discriminatory responses
against certain individual users or user groups, leading to the spread
of biases and stereotypes and causing harm to the society. While the
bias can sometimes be unintentional and arise from a variety of factors,
fairness is an intentional goal that people strive to achieve. To judge
whether a system is fair, we need to define fairness first. It is,
however, difficult to give a universal definition of fairness.
Fairness is a complex concept that depends on the application
context. Different groups and individuals see fairness differently.
Multiple definitions of fairness are discussed in
\cite{verma2018fairness, mitchell2021algorithmic, caton2020fairness,
mehrabi2021survey}.  Each of these definitions represents unique
perspectives on the applications and the interests of different groups. 

Most fairness definitions can be roughly divided into four
categories: group fairness, individual fairness, causal fairness, and
counterfactual fairness. They are elaborated below.
\begin{itemize} 
\item Group fairness, such as equal opportunity and demographic parity,
focuses on differences between the chatbot's responses to different
groups given similar prompts. It aims to eliminate discrimination
against certain groups. For example, when a chatbot is asked to
create jokes related to a specific racial group using stereotypes, it
should reject such requests consistently across all the groups. Group
fairness cannot be achieved if requests regarding certain racial groups
are denied while requests regarding other racial groups are
accommodated.
\item Individual fairness, such as fairness through awareness,
emphasizes differences between responses received by individuals of
similar backgrounds. For example, if a chatbot is asked to
provide rehabilitation recommendations, patients with
similar conditions should receive similar recommendations rather than
completely different recommendations based on demographics such as
gender or age. If a male patient is advised to exercise more and eat
more protein, while a female patient is advised to rest more, the
chatbot does not meet the goals of individual fairness.
\item Causal fairness evaluates the fairness of a system from the
perspective of changing a specific characteristic and observing biases
in the response \cite{galhotra2017fairness}. It aims to mitigate
the impact of specific attributes on decisions and avoid the system
perpetuating historical biases and inequalities. For example, if a
chatbot is asked to recommend products, it should give balanced
suggestions to all users based on individual preferences. If the chatbot
primarily recommends cosmetics to women and electronics to men only
based on historical data, it may propagate historical gender-based
inequalities.
\item Counterfactual fairness \cite{kusner2017counterfactual}
evaluates fairness by considering hypothetical scenarios where
sensitive attributes are different while other attributes are the same.
For example, if someone asks for chatbot assistance with a hiring
decision. The person could then create a counterfactual scenario by
switching the gender of the applicant to see whether the advice given by
the chatbot is consistent. If they are inconsistent, counterfactual
fairness is not met.
\end{itemize} 

Each fairness definition is reasonable. However, satisfying all of them at the
same time is challenging \cite{kleinberg2016inherent}. To design a fair
chatbot, it is crucial to clarify its application context, including its
purpose, who will use it, how it will be used, its difference from
traditional methods, people's attitudes, and possible loopholes. In the
development process, designers should consider additional questions,
e.g., what sensitive features may be implied in the prompt words, which
fairness definitions should be selected and their priority, whether
current fairness considerations will change over time, what causes the
differences between groups or individuals and whether they are
reasonable, and so on. In addition, designers should think about the
consequences of false positives and false negatives.  When a chatbot
makes mistakes, which of the two will have a more serious outcome?  For
example, for a chatbot used for children's education, the impact of not
blocking inappropriate content by mistake is much worse than blocking
irrelevant content by mistake. 

\section{Future Research Directions}\label{sec: Future}

\subsection{Open-domain Versus Domain-specific Chatbot}

Open-domain chatbots, such as ChatGPT, have demonstrated their power
recently. They can handle prompts in multiple domains of complex social
contexts and from a wide range of user groups. Generally speaking, it is
difficult to mitigate biases and develop fair open-domain chatbots.
Domain-specific chatbots are different. They have specific user groups
and preset application scenarios. Thus, it is easier to consider
possible biases, choose the appropriate fairness metrics, and implement
a relatively fair system. Besides, open-domain chatbots with LLMs
usually require a huge amount of training data and computing resources.
While domain-specific chatbots have limited knowledge, they demand much
less training data and computing resources. They are easier to control.
In some fields where accuracy or user privacy is important, such as
healthcare, domain-specific chatbots are more likely to obtain accurate
responses than open-domain chatbots under the same amount of resources.
It is also easier to protect users' privacy in domain-specific chatbots
with a smaller model size that runs locally without uploading data to the
public server. 

\subsection{Bias Control in Multi-Modal Chatbots}

Chatbots with multi-modal input/output will become the main trend in the
future. Such chatbots not only need to take care of NLP and dialogue
management, but they also need multiple models for modality conversion and
integration. To realize a fair chatbot system, modality conversion and
integration models have to satisfy their respective fairness metrics.
After putting them together in the system, they may affect each other
and the overall output may be biased. It is important but challenging to
mitigate the bias of the whole system. 
        
\subsection{Green and Interpretable Chatbots}

LLM-based chatbots become popular recently. Since LLMs need a huge
amount of computing resources to train, they are not environment
friendly.  The chatbots are built upon large pre-trained models with
fine-tuning. They contain biases from multiple sources. They are
difficult to mitigate since developers treat the whole system as a black
box. To detect and mitigate biases, one can only rely on input prompts
and output responses. As a result, the bias detection job is labor
intensive. On the one hand, traditional chatbots contain less bias.  On
the other hand, they are inferior to LLM-based chatbots in generating
fluent human natural language with rich and diversified content.  It is
appealing to design a logically transparent yet content-rich chatbot. 

One possible direction is to leverage the tool of knowledge graphs
(KGs).  The knowledge graph (KG) provides an efficient and clear data
structure to store human knowledge.  It can be used for knowledge
reasoning and retrieval.  There have been efforts to enhance the
knowledge base of LLMs with KGs \cite{sun2023think, pan2023unifying}.
However, it still cannot offer a transparent reasoning process. 

With the rapid increase in the size of ML models and the required
training resources, green AI \cite{schwartz2020green, kuo2022green} has
gradually gained attention.  Green AI technologies may have the
potential to be used in developing chatbots with clearer reasoning
processes, smaller model sizes, fewer training resources, and more
environmental friendliness.  The green learning (GL) methodology
\cite{kuo2022green} has been shown to offer comparable performance with
Deep Learning (DL) in many applications. GL methods have much smaller
model sizes and lower inference complexities (in terms of FLOPs). They
also demand fewer training samples. The GL-based chatbot design may lie
in the decomposition of LLMs into two modules: 1) GL-based language
models as the interface with respect to users for NLU and NLG tasks; and
2) GL-based KGs as the core for knowledge storage, expansion, search,
and reasoning. This high-level idea may guide us to develop a more
transparent and scalable chatbot. Biases in GL-based chatbots can be
traced, making the implementation of a fair system easier. 

\section{Conclusion}\label{sec: Conclusion}

As an AI system that communicates directly with humans, chatbots have a
long development history. They have received special attention in recent
years due to the amazing performance of LLM-based chatbots, such as
ChatGPT.  Although traditional chatbots are relatively rigid with
limited functionalities, their models are smaller and easier to deploy.
They have fewer bias and discrimination concerns. They are suitable for
small-scale domain-specific applications.  Chatbots have become much
more powerful with the rapid advancement of LLMs and computing resources in
recent years. On the other hand, they have brought controversy about
bias and fairness.  From development teams to users, every human
interacting with an ML-based chatbot has the potential to spread the
bias.  To design and deploy a fair chatbot system, the development team
needs to know social needs, design goals, and actual effects. Overall, the
bias and fairness issues of chatbots remain an open problem
demanding further research.

\bibliographystyle{plain}
\renewcommand\refname{Reference}
\bibliography{main}

\end{document}